\documentclass[runningheads]{llncs}
\usepackage{graphicx}
\usepackage{xcolor}
\usepackage{amsmath}
\usepackage{caption}
\usepackage{array}
\usepackage{url}
\usepackage{textcomp}
\usepackage{booktabs}

\newcommand{\squishlist}{
 \begin{list}{$\bullet$}
  { \setlength{\itemsep}{0pt}
     \setlength{\parsep}{5pt}
     \setlength{\topsep}{5pt}
     \setlength{\partopsep}{0pt}
     \setlength{\leftmargin}{1.5em}
     \setlength{\labelwidth}{1em}
     \setlength{\labelsep}{0.5em} } }
 \newcommand{\squishend}{\end{list}}

\DeclareCaptionType{exam}[Example][List of Examples]

\begin{document}
\title{Enriching Relation Extraction with OpenIE}
%
%
\author{Alessandro Temperoni\inst{1} \and
Maria Biryukov\inst{1} \and
Martin Theobald\inst{1}}
\authorrunning{A. Temperoni et al.}
%
\institute{University of Luxembourg, 2 Avenue de l'Universite, Esch-sur-Alzette, Luxembourg} 
\maketitle              
\begin{abstract}
Relation extraction (RE) is a sub-discipline of information extraction (IE) which focuses on the prediction of a relational predicate from a natural-language input unit (such as a sentence, a clause, or even a short paragraph consisting of multiple sentences and/or clauses). Together with named-entity recognition (NER) and disambiguation (NED), RE forms the basis for many advanced IE tasks such as knowledge-base (KB) population and verification. In this work, we explore how recent approaches for open information extraction (OpenIE) may help to improve the task of RE by encoding structured information about the sentences' principal units, such as subjects, objects, verbal phrases, and adverbials, into various forms of vectorized (and hence unstructured) representations of the sentences. Our main conjecture is that the decomposition of long and possibly convoluted sentences into multiple smaller clauses via OpenIE even helps to fine-tune context-sensitive language models such as BERT (and its plethora of variants) for RE. Our experiments over two annotated corpora, KnowledgeNet and FewRel, demonstrate the improved accuracy of our enriched models compared to existing RE approaches. Our best results reach 92\% and 71\% of F1 score for KnowledgeNet and FewRel, respectively, proving the effectiveness of our approach on competitive benchmarks.
\end{abstract} 


\section{Introduction} 
\label{sec:introduction}

Relation extraction (RE) is a way of structuring natural-language text by means of detecting potential semantic connections between two or more real-world concepts, usually coined ``entities''. Relations are assumed to fall into predefined categories and to hold between entities of specific types. For example, the {\sc SPOUSE} relation may exist between two entities of type ``Person'', while instances of the {\sc CEO} relation would link entities of type ``Person'' and ``Organisation'', respectively. Being itself a sub-discipline of information extraction (IE), extracting labeled relations may also help to boost the performance of various IE downstream tasks, such as knowledge-base population (KBP) \cite{Weikum:2019,Gardner:2015} and question answering (QA) \cite{Wang:2012,XU:2016}.

\smallskip\noindent\textbf{Distant Supervision vs. Few-Shot Learning.~} Extracting labeled relations from previously unseen domains usually requires large amounts of training data. Manually annotated corpora are relatively small due to the amount of work involved in their construction. To this end, {\em distant supervision} \cite{Mintz:2009} may help to alleviate the manual labeling effort but training data, which may serve as the basis for distant supervision, is only available for relations covered by an already-existing KB (such as Yago~\cite{suchanek2007yago}, DBpedia~\cite{dbpedia} or Wikidata~\cite{vrandevcic2014wikidata}).
To overcome this limitation, \cite{gashteovski} manually evaluates the semantics of alignments between OpenIE triples and the facts in DBpedia. There, distant supervision is used as a first step to compare facts that have the same (or at least similar) arguments. We, in contrast, use distant supervision to transfer the labels from the annotated corpora to the OpenIE extractions, thereby creating an annotated set of clauses which can then be used for training. Moreover, for cold-start KBP settings \cite{KBP-Cold-Start-Track}, {\em few-shot learning} \cite{fewshot} has recently evolved as an interesting alternative to distant supervision. In few-shot-training for KBP (or more classical tasks like text classification), an underlying language model such as BERT or SBERT \cite{Devlin:2018,SBERT} is augmented by an additional prediction layer for the given labeling task which is then retrained by very few samples. Here, often 20--50 examples for each label are sufficient to achieve decent results. However, all of these approaches for KBP focus on labeling and training trade-offs for the given input text, while other--perhaps more obvious---options, namely to exploit syntactic and other structural clues based on OpenIE, NER and NED, are at least as promising as these training aspects in order to further improve prediction accuracy.

\smallskip\noindent\textbf{Domain-Oriented vs. Open Information Extraction.~} OpenIE \cite{NELL,KnowItAll,Banco:2007,Fader:2011} expresses an alternative text-structuring paradigm compared to the more classical, domain-oriented IE techniques  \cite{Weikum:2019,Gardner:2015}: it transforms sentences into a set of {\textit{\textlangle{}{arguments -- relational phrase\textrangle{}}} tuples without labeling the relational phrases explicitly or requiring its arguments to be of particular entity types. Consider, for instance, the sentence: \textit{``In 2008 Bridget Harrison married Dimitri Doganis''}. From an RE perspective, it would be represented as: \textlangle{}Bridget Harrison; SPOUSE; Dimitri Doganis\textrangle{}. Its OpenIE\footnote{Based on OpenIE 5.1: \url{https://github.com/dair-iitd/OpenIE-standalone}} counterpart would decompose the input sentence into two tuples: \textlangle{}Bridget Harrison; married; Dimitri Doganis\textrangle{} and \textlangle{}Bridget Harrison; married Dimitri Doganis In; 2008\textrangle{}. Intuitively, the two representations capture the same semantic message of a marriage relationship between Bridget Harrison and Dimitri Doganis. Furthermore, OpenIE produces additional informative tuples describing, e.g., temporal or collocational aspects of the relation via adverbial phrases, which however may not necessarily have a corresponding canonicalized form. Importantly, OpenIE extracts relational phrases along with the original sentences' arguments, thus structuring the input text without loss of information. All these characteristics make OpenIE a useful intermediate representation for a number of downstream IE tasks that impose further structuring or normalization \cite{Mausam:2016,Martinez-Rodriguez2018,OpenCeres:2019}. 

\smallskip\noindent\textbf{Word Embeddings vs. Language Models.~} In the past few years, {\em word embeddings} \cite{Bojanowski:2017,Mikolov:2013,glove} found their applications and proved to be efficient in a wide range of IE tasks. Word embeddings represent text as dense vectors in a continuous vector space. Traditional word embeddings, such as Word2Vec \cite{Mikolov:2013} and FastText \cite{Bojanowski:2017}, are lightweight and conveniently fast at training and inference times. However, being static (in the sense that each word in the corpus is always represented by the same vector, regardless of its context), these embeddings have a very limited ability to capture a word's changing meaning with respect to different contexts. 
On the contrary, recently trained, large-scale {\em language models} (LMs), such as BERT \cite{Devlin:2018}, ELMO \cite{Peters:2018} or GPT-3 \cite{Radford:2019}, extend the approach by generating dynamic embeddings, where each word's representation depends on its surrounding context, thus pinning down particular meanings of polysemic words and entire phrases. Despite the differences, both types of embeddings allow to quantitatively express semantic similarities between words and phrases based on the closeness of their respective vectors in the vector space. Furthermore, besides their purely lexical input, other linguistic components such as syntactic dependency trees or OpenIE-style tuples can be used to train or fine-tune various embedding models with positive impact on more advanced IE tasks such as text comprehension, similarity and analogy \cite{Levy:2014,Stanovsky:2015}, semantic role labeling (SRL) \cite{srl}, as well as RE and QA \cite{Sachan:2021}.   

\smallskip\noindent\textbf{Contributions.~} 
In this work, we systematically investigate various combinations of the above outlined design choices for the task of RE. Specifically, we combine OpenIE with both types of embeddings (i.e., context-free and context-sensitive ones) and examine the strengths and limitations of each combination. Our main conjecture is that OpenIE is able to improve even context-sensitive LMs such as BERT because it decomposes potentially large sentences into multiple clauses which each represent the target relation in a sharper manner than the original sentence. We summarize our motivation for investigating a combination of OpenIE and LMs for the task of RE as follows.
\squishlist
\item Our goal is to advance Web-scale relation extraction. To this end, we adopt the OpenIE approach to model and classify relational phrases by leveraging shorter clauses which more accurately capture the target relation than potentially long and convoluted input sentences.
\item We transfer the labels from the annotated corpora to the OpenIE extractions in a distant-supervision fashion, thereby limiting the manual labeling effort that is otherwise needed for training and fine-tuning the underlying models. We also systematically investigate few-short training which is able to further reduce the amount of labeled training examples to less than 20 per relation (and yet yield very satisfactory results in many cases).
\item We perform detailed experiments on two annotated RE corpora, namely \textit{KnowledgeNet} \cite{KN} and \textit{FewRel} \cite{FewRel1}, using Wikidata as a backend KB in combination with various state-of-the-art (both context-free and context-sensitive) LMs: a basic Word2Vec model with both annotated and disambiguated NEs, BERT, RoBERTa, AlBERT, SETFIT, plus their ``distilled'' versions. Various of our combined approaches are able to improve over the best known results for both KnowledgeNet and FewRel by partly very significant margins.
\squishend
 
The rest of the paper is organized as follows: we present our general methodology in Section \ref{sec:meto}, introduce our pipeline in Section \ref{sec:pipeline}, describe the experimental setup in Section \ref{sec:experimental}, and show the results in Section \ref{sec:evaluation}. 
\section{Methodology}
\label{sec:meto}
In this section, we present our three principal strategies for classifying relational paraphrases (and entire clauses) obtained from OpenIE into canonical relations over a predefined KB schema. We next provide a brief overview of the three approaches, before we describe them in more detail in the following subsections.
\vspace*{-7mm}
\begin{figure*}[!ht]
\centering
\includegraphics[width=1.3\textwidth]{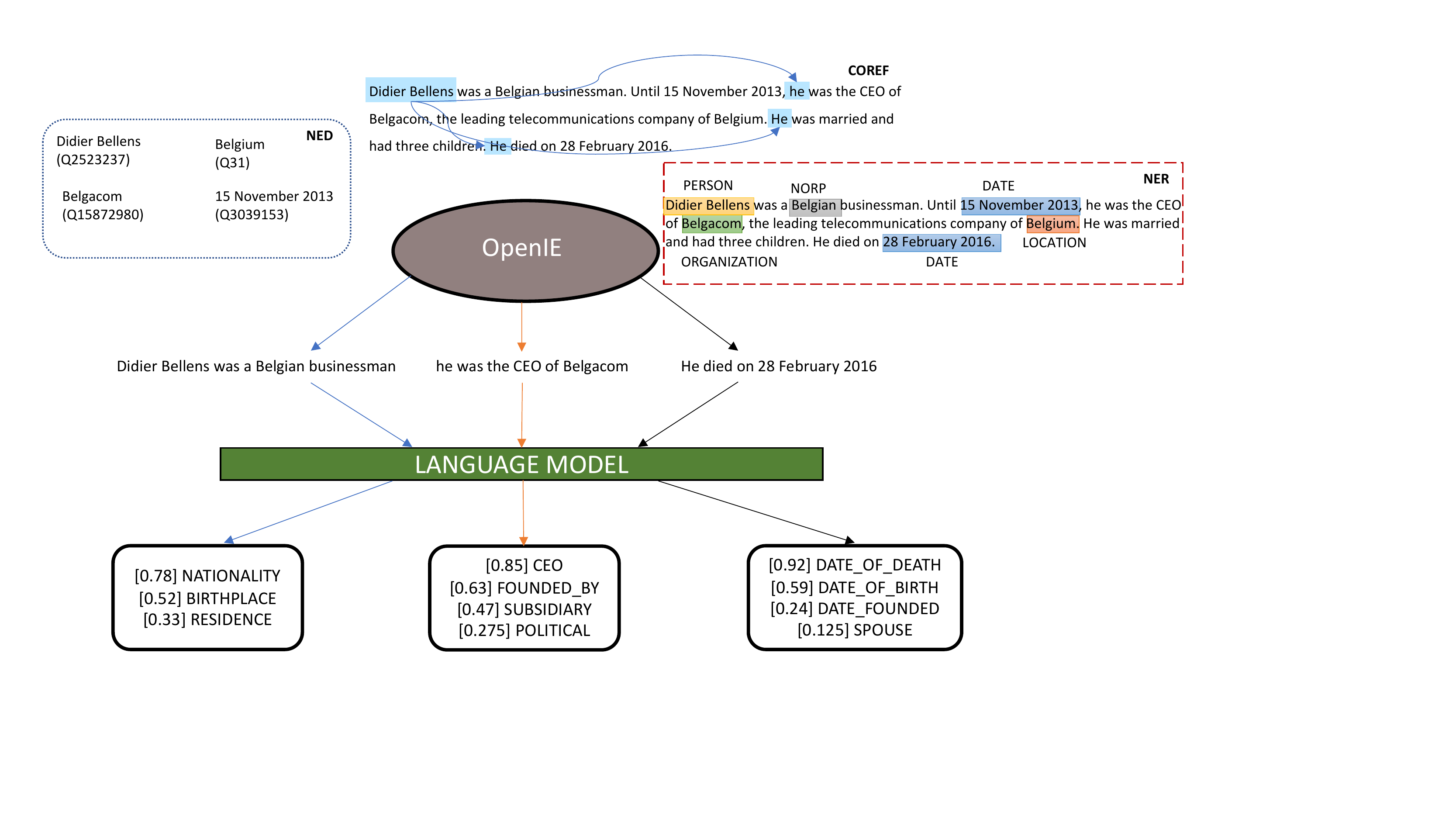}
\caption{System overview.}
\label{fig:architecture}
\end{figure*}
\vspace*{-7mm}

\smallskip\noindent\textbf{Fine-Tuning Language Models.~} Our first approach (see Subsection \ref{sub:bert} for details) to achieve this goal is to train a dedicated RE model from a corpus of annotated sentences and to then use this RE model to predict the relations for previously unseen sentences and clauses. Specifically, we start with a large-scale, pretrained LM, such as BERT (or one of its variants), and add a classification layer on top in order to {\em fine-tune the model} on the RE classification task. As BERT is a general-purpose, context-sensitive LM trained on many billions of input sentences, we expect this approach to work best for RE, with just a small amount of annotated sentences being required for fine-tuning the classification layers.

\begin{figure}[!ht]
\centering
\begin{minipage}{.5\textwidth}
\centering
\includegraphics[width=.9\linewidth]{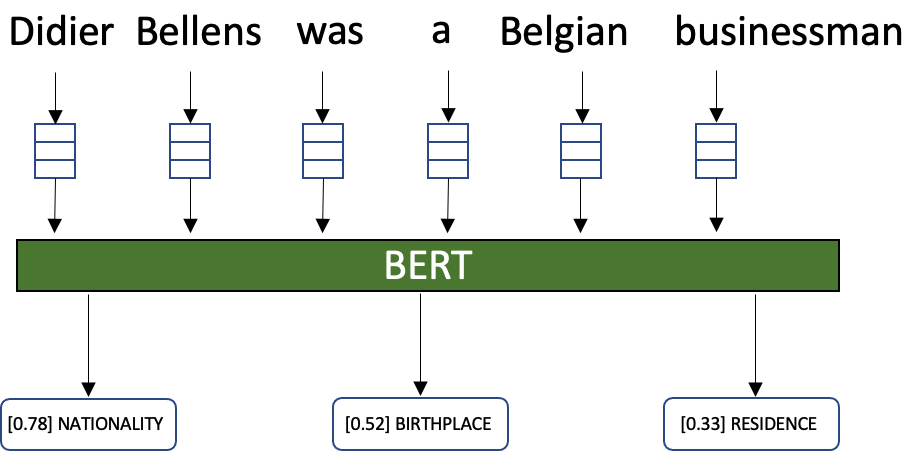}
\label{fig:example1}
\end{minipage}%
\begin{minipage}{.5\textwidth}
\centering
\includegraphics[width=.9\linewidth]{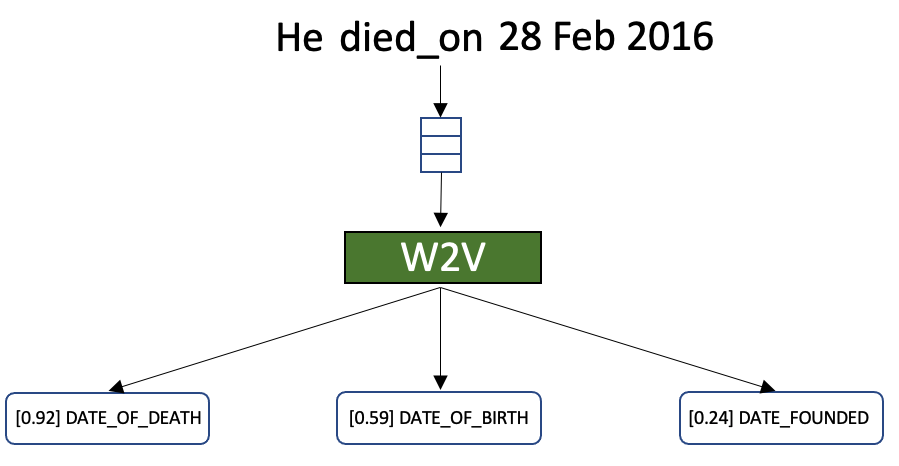}
\label{fig:example2}
\end{minipage}
\end{figure}

\smallskip\noindent\textbf{Context-Free Relation Signatures.~} As a simpler, context-free baseline to the above approach (see Subsection \ref{sub:corpus2triplet}), we also investigate the usage of a clause-based Word2Vec model for RE, which requires no annotated sentences (at least for training) at all. Here, we directly train the Word2Vec model over a domain-specific corpus (such as Wikipedia articles) in an unsupervised manner. By aggregating individual word vectors into {\em relation signatures} for a given set of target relations, we quantitatively assess the vector similarities between these relation signatures and the relational paraphrases obtained by OpenIE. 

\smallskip\noindent\textbf{Contextualized Relation Signatures.~} Our third and final approach (see Subsection \ref{sub:relsignatures}) combines the above two ideas by investigating the usage of BERT-like models in a feature-based manner. Here, the contextualized embeddings extracted from a large-scale pre-trained model constitute an input for a {\em contextualized form of relation signatures} by manually providing a few training sentences as input for each such relation signature.

\subsection{Fine-Tuning Language Models for Relation Extraction}
\label{sub:bert}
In the context-aware approach, we add a single fully connected layer for the classification task on top of the last layer of an otherwise task-agnostic pre-trained LM such as BERT or one of its variants (see below for details). The size of the added layer is equal to the number of classification labels. Fine-tuning the model then consists of training the new layer's weights over a task-specific annotated dataset. For our RE task, a typical annotated example would consist of (1) an {\em input sentence}, (2) the {\em entity pair} corresponding to the sentence's subject and object, and (3) the {\em target relation} as label. 

For example, the sentence \emph{ "After five successful albums and extensive touring, they disbanded after lead vocalist Sandman died of a heart attack onstage in Palestrina, Italy, on July 3, 1999."} would then be encoded into the clause (amongst others) \textlangle{} \{Sandman; July 3, 1999\}, {\sc DATE\_OF\_DEATH} \textrangle{}. Note, however, that our approach to the relation classification differs from the established setup in one important way: while many works on the topic capitalize on the importance of relational argument (entity) representation \cite{SoaresFLK19,Zhou:2021,Boros:2021,OpenKI}, we completely exclude entity-related information (obtained from common NER/NED toolkits) during the training, thereby delegating the task of extracting the relational argument to the OpenIE step. Therefore, an adjusted input for tuning the model is reduced to pairs made of (1) {\em input clause} and (2) {\em target relation}.

\smallskip
The BERT family of LMs we used for fine-tuning is listed below. We briefly introduce each model and motivate our choices.  
\squishlist
\item \emph{bert-base-uncased} is a \textbf{B}idirectional \textbf{E}ncoder \textbf{R}epresentations from \textbf{T}ransfor-mers (BERT) model \cite{Devlin:2018}, which consists of 12 layers of transformers and is case-insensitive. BERT became a default ``baseline'' for many NLP tasks involving general-purpose pre-trained models. 
\item \emph{distilbert-base-uncased} \cite{distilBert:2020} is a variant of BERT, pre-trained on the knowledge distillation principle which consists of transferring knowledge from (a set of) large model(s) to a single smaller one. DistilBERT has been shown to perform nearly on par with the full-size BERT while using less resources and being faster at training and inference time. 
\item \emph{xlnet-base-cased} \cite{XlNET:2019} is an autoregressive model that improves on BERT's capability of learning semantic dependencies between sentence components. This feature may have an advantage at inference time when we use a model that is fine-tuned on entire sentences in order to classify OpenIE-style clauses.   
\item \emph{roberta-base} \cite{RoBERTa:2019} has been trained on a much larger corpus than BERT (and is yet optimised). It also introduces a dynamic token-masking objective to the model's training in order to increase its sensitivity and robustness to the context. It showed the highest accuracy (compared to BERT and XLNet) on the task of Recognizing Textual Entailment (RTE) \cite{RoBERTa:2019,GLUE:2018} which is closely related to the task of RE. This motivates our interest in using this model.
\item \emph{albert-base-v1} \cite{ALBERT:2019} finally introduces a sentence-order prediction (SOP) training objective which focuses primarily on inter-sentence coherence---a property we expect to leverage from when transferring knowledge learned from entire sentences to OpenIE clauses. 
\item \emph{setfit} \cite{setfit} stands for \textbf{Se}ntence \textbf{T}ransformer \textbf{Fi}ne-\textbf{t}uning and is a very recent model designed for few-shot text classification. It is trained on a small number (8, to be exact) of text pairs in a contrastive Siamese manner. The resulting model is then used to generate rich text embeddings which are used to train a classification task.

\squishend

\subsection{Using Context-Free Relation Signatures for Relation Extraction}
\label{sub:corpus2triplet}

For the context-free approach, we start from a large dump of English Wikipedia articles which we process with a pipeline consisting of ClausIE \cite{CIE} for clause decomposition, Stanford CoreNLP \cite{manning2014stanford} and AIDA-light \cite{aida} for named-entity recognition (NER) and -disambiguation (NED), respectively. This pipeline yields an initial amount of 190 million clauses, from which we distill 13.5M binary relations of the form {\em\textlangle{}{subject; relational phrase; object\textrangle{}}} (thereby keeping only clauses that have exactly one named entity both as subject and object as well as a verbal phrase as predicate).

Following \cite{Fader:2011}, we apply regular expressions on the verbal phrases to identify patterns of the form \emph{verb} $\vert$ \emph{verb + particle} which should cover $\approx 85\%$ of the verb-based relations in English. To further normalize the extracted verbal phrases, we use part-of-speech (POS) and lemmatization information: for all but passive verbs, we substitute their inflections with the respective lemma. This way, we are able to distinguish between active and passive voices of otherwise identical verbs (which usually indicate inverse relations of each other). After the above cleaning and normalization steps, our overall representation of a clause is of the form: \textlangle{}$\mathit{entity}_1$, $\mathit{verb} + \mathit{particle}$, $\mathit{entity}_2$ \textrangle{} with the additional condition that $\mathit{entity}_1$ and $\mathit{entity}_2$ should not be equal\footnote{We found about 5\% of clauses in the corpus to represent reflexive relations, i.e., with the subject and object referring to the same entity. From our observations, we could not derive meaningful KB facts from these relations. We therefore removed such reflexive relations from the corpus.}. 

We next embed the clauses into their word vector representations. Specifically, we consider two encoding schemes: 
\begin{itemize}
    \item[(i)] by exploiting the {\em compositionality} of word vectors:
 $$ \vec{V}_\mathit{verb} + \vec{V}_\mathit{particle} $$
    \item[(ii)] by creating {\em bigrams} of verbs and particles for the most frequent relational paraphrases in the corpus (e.g., \emph{work\_at, graduate\_from, born\_in}):
$$ \vec{V}_{\mathit{verb}\_\mathit{particle}} $$
\end{itemize}
For the latter bigram-based encoding, we treat bigrams for the prepositional verbs as additional dictionary entries before a Word2Vec model is trained on the clauses. This choice is motivated by considering that, particularly for knowledge discovery, particles (i.e., prepositions) may give a crucial insight on the possible type of the entities involved in the relation. As we will see in Section \ref{sec:experimental}, we leverage both aforementioned techniques in comparison.

To train the models under (i) and (ii), we use 
Word2Vec ``skip-gram model'' implementation provided by the Gensim 
\cite{rehurek2011gensim}, with the window size 2, and negative sampling as loss function.  

In this context-free approach, we further aggregate the vector representation of each target relation by including also synonyms for these relations provided by an additional backend KB. As an example, let 
$$\mathit{S}_{\mathtt{P571}} =  \{\text{``date founded''}, \text{``date created''}, \ldots, \text{``established''}\}$$
 denote the {\em Wikidata\footnote{\url{www.wikidata.org}} synonyms} provided for the relation $\mathtt{P571}$. Then, the vector for its corresponding {\em relation signature} is computed as follows
 $$\vec{V}_{\mathtt{P571}} ~=~ \frac{1}{|\mathit{S}_{\mathtt{P571}}|} ~ \sum_{\mathit{synonym} \in \mathit{S}_{\mathtt{P571}}}{\vec{V}_{\mathit{synonym}}}$$
where we use the {\em artihmetic mean} (in analogy to Gensim\footnote{\url{https://radimrehurek.com/gensim/}}'s {\tt w2v.most\_similar} function to retrieve similar vectors for a set of positive examples) in order to aggregate a set of such synonyms into a single vector.

Since the target relations considered in our experiments correspond to Wikidata \cite{vrandevcic2014wikidata} properties, we use Wikidata as backend KB and consider the English parts of the \emph{``Also known as''} sections of the respective properties as source for the synonymous relational phrases. To leverage our Word2Vec model, we again normalise the property name and its synonyms by following the steps described above (before vectorization). By default, we then use bigrams of verb lemmas and their particles for the aggregation of the vectors into relation signatures. However, if a bigram is not found in the model's vocabulary, we fall back to our compositional encoding also for the respective synonyms.  

\subsection{Using Contextualized Relation Signatures for Relation Extraction}
\label{sub:relsignatures}

For our third approach, we further build on the idea of using relation signatures to represent relations but this time generate {\em contextualized relation signatures} in a slightly different way. Since the \emph{``Also known as''} sections of the Wikidata properties contain just plain lists of phrases representing the given properties, they are not directly suited for the BERT-based LMs we discussed earlier. We therefore randomly sample 5 different sentences in the underlying corpus for each target property manually in order to fine-tune the LMs. These sentences are then removed from the evaluation set.
This heuristic is purposely implemented to resemble few-shot learning techniques in a feature-based manner. It has two major advantages for our goal of scaling RE: it involves a very low amount of additional labeling effort, and it allows to add new target relations on-the-fly.
\section{Text-Processing Pipeline}
\label{sec:pipeline}
Before we apply the methods described above to an actual corpus, we first preprocess the corpus with a typical NLP pipeline consisting of $5$ steps: sentence splitting, tokenization and part of speech tagging\footnote{\url{https://spacy.io/}}, NER \cite{FlairT2021} and NED  \cite{ELQ} (both optional\footnote{In the scope of this work we focus on the relational predicates and use named entity recognition (NER) to ensure that the subject and object entity types are compatible with the backend KB property.}), coreference resolution (CR) \cite{JoshiSB} and OpenIE \cite{Mausam:2016}.
Once all the steps are performed, we align the obtained annotations with the original sentences to obtain a corpus representation that is of the form shown in Figure \ref{fig:architecture}. Clauses, now enriched with the POS, NER/NED and CR information, serve as input for the RE task.  

\subsection{Model Tuning Strategy}
\label{sub:tuning}

Since we aim at leveraging OpenIE for RE, a natural choice for tuning a language model would be to use a set of labeled clauses as input, where the labels correspond to the target relations for a given dataset. To the best of our knowledge, such datasets are not available at clause level. We therefore create one in a distant-supervision manner as follows: given a sentence from the training corpus with the relation specified as a pair of \textlangle{}{\em subject}, {\em object}\textrangle{} entities and a label, we label the clauses obtained from that sentence provided that their subject and object correspond to the subject and object entities from the labeled sentence. 

\vspace{8px}

\quad \quad PLACE\_OF\_BIRTH(\underline{Barack Obama}, \underline{Honolulu})

\vspace{1px}

\quad \quad OpenIE clause: \textit{\underline{Barack Obama} was born in \underline{Honolulu}}

\vspace{1px}

\quad \quad (\textit{\underline{Barack Obama}  was born in \underline{Honolulu}}, PLACE\_OF\_BIRTH)

\vspace{8px}

Note that we can only safely align clauses to labeled relations when the former correspond to unambiguous extractions with exactly one entity in the subject and object, respectively. 

It is generally believed that a clause's predicate is the main carrier of the relation type \cite{Fader:2011,CompactIE:2022,OpenCeres:2019}. We observe though that it is not necessarily the case. Consider the following clauses: (a) \textlangle{}John Deane Spence; was; a \textit{British Conservative Party politician}\textrangle{} and (b) \textlangle{}Eccles; served as; a \textit{Labour Party member} of Manchester City Council\textrangle{}. In these examples, the clause's object contains both -- the cue of the relation type (\emph{Political Affiliation}) and the relational argument. In other cases, the predicate contains both -- a relation type indicator and relational argument: (c)\textlangle{} he; \textit{joined the Communist Party} In; 1923.\textrangle{} Note that the clause's object here is irrelevant for the relation. These examples show that there is no ``consistency'' in the OpenIE format that we could rely on when translating OpenIE extractions to RE. Gashteovski\cite{Gemulla:2017} addresses the problem of clause normalization by applying a series of transformations driven by a morpho-syntactic analysis of the produced clauses. 

On the contrary, we consider the clauses produced by an OpenIE system as potentially noisy relations. 
This is the reason why we consider each OpenIE extraction in its entirety, as a short sentence, and use each for the relation prediction individually. In this way, we exploit an implicit semantic connection between the clause's elements that synergically express the relational meaning of a clause. As such our approach does not suffer from falling into one of the two extremes as indicated in \cite{OpenKI}: neither performing an instance-level inference relying on embedding for \textlangle{}{\em subject}, {\em object}\textrangle{} pairs thus being unable to handle unseen entities; nor performing a predicate-level mapping thus ignoring background evidence from individual entities. Rather, we examine to what extent task-agnostic pretrained LMs are able to transfer learned signals from longer sentences to short facts. We therefore use a portion of labeled sentences to tune the model and use it for clause classification.     
\section{Experiments}
\label{sec:experimental}
In this section, we describe the experiments and datasets we used to evaluate our proposed methods, which are based on two commonly used RE benchmarks: \emph{KnowledgeNet} and \emph{FewRel}. Both collections were preprocessed by our NLP pipeline described in Section \ref{sec:pipeline}.
  

\smallskip
\noindent\textbf{KnowledgeNet} (KN) \cite{KN} is a dataset for populating a KB (here: Wikidata) with facts expressed in natural language on the Web. We selected KN as our primary benchmark because it provides facts in the form of \textlangle{}{\em subject}, \emph{property}, {\em object}\textrangle{} triplets as sentence labels. The documents in the first release of KN are either DBpedia abstracts (i.e., the first paragraph of a Wikipedia page) or short biographical texts about a person or organization from the Web. 
9,073 sentences from 4,991 documents were chosen to be annotated with facts corresponding to 15 properties  
(see \cite{KN} for detailed list).
In total, KN comprises 13,425 facts from 15 properties. Only 23\% of the facts annotated in their release are actually present in Wikidata. 

\smallskip
\noindent\textbf{FewRel} \cite{FewRel1} is a very popular benchmark for few-shot RE, consisting of 70,000 sentences over 100 relations (divided in 64 for training and 36 for testing purposes). This dataset is meant to be very competitive even for the most advanced models for RE, and for this reason we employed it also in our work. However, we did not use FewRel as it was originally conceived in its typical few-shot setting, but we randomly split the sentences per relation into separate training (75\%) and testing (25\%) sets.

\subsection{Baseline Approaches}
\label{sec:baselines}
We now evaluate the three different approaches of Section \ref{sec:meto}. Particularly, for the fine-tuned BERT models, we created different combinations of training and testing sets as follows.
\squishlist
\item \textbf{Baseline 1: Clauses + LM.~} We use OpenIE to extract clauses from sentences. Both fine-tuning and prediction of the LM were then performed on {\em clauses}.
\item \textbf{Baseline 2: Mixed + LM.~} Fine-tuning of the LM was performed on {\em sentences}, while prediction was then performed on {\em clauses}.
\item \textbf{Baseline 3: Sentences + LM.~} Both fine-tuning of the LM and inference were performed on {\em sentences}.
\item \textbf{Baseline 4: Clauses + W2V.~} We use OpenIE to extract clauses from sentences. Context-free relation signatures (as described in Section~\ref{sub:corpus2triplet}) based on the simple Word2Vec model were then used to infer the target relation. 
\item \textbf{Baseline 5: Clauses + feature-based BERT}. For the feature-based approaches, we applied the same three combinations as for Baselines 1, 2, and 3. Thus swapping the fine-tuning phase with the \textit{relation signature} construction which was generated by using only 5 randomly drawn samples per relation. For this baseline, both \textit{relation signature} construction and inference were performed on {\em clauses}.
\item \textbf{Baseline 6: Mixed + feature-based BERT}. The \textit{relation signature} construction was generated using 5 randomly drawn {\em sentences} per relation, while prediction was performed on {\em clauses}. 
\item \textbf{Baseline 7: Sentences + feature-based BERT}. Both the \textit{relation signature} construction and prediction were performed on {\em sentences}.
\squishend 
\subsection{Evaluation}
\label{sec:evaluation}
RE inherently resembles a {\em multi-class prediction} task. For KN, a particularity of the benchmark is that sentences may also have multiple labels, i.e., we need to consider and evaluate a {\em multi-label prediction} setting. Moreover, since OpenIE may turn each input sentence into multiple clauses, we define the following variants of the three classes of {\em true positives} (TPs), {\em false positives} (FPs) and {\em false negatives} (FNs) needed to compute precision, recall and F1, and with respect to whether the unit of prediction is either a {\em sentence} or a {\em clause}.

\smallskip
\noindent\textbf{Prediction Unit: Sentence.~} The unit of prediction is a sentence. Under a {\em single-label prediction} setting, TPs, FPs and FNs can be computed in the standard way by considering also a single (i.e., the ``best'') predicted label per sentence. However, under a {\em multi-label prediction} setting, we predict as many labels as were given for the KN sentence, and then consider how many of the predicted labels also match the given labels as the TPs (and vice versa for the FPs and FNs).

\smallskip
\noindent\textbf{Prediction Unit: Clause.~} The unit of prediction is a clause. Under a {\em single-label prediction} setting, this means that we also predict one label per clause, but since OpenIE may extract multiple clauses from the given KN sentence, we then still need to compare multiple labels obtained from the clauses with the single, given label of the KN sentence. We therefore define the following two variants for TPs and FPs (FNs again follow similarly): ANY and ALL.

\medskip\noindent
\begin{tabular}{lll}
\textbf{ANY} & ~TP:~ & {\em any} of the clauses' labels match the single given label of the KN\\
& & sentence.\\
& ~FP:~ & {\em none} of the clauses' labels match the single given label of the KN\\
& & sentence.\\
\textbf{ALL} & ~TP:~ & {\em all} of the clauses' labels match the single given label of the KN\\
& & sentence.\\
& ~FP:~ & {\em not all} of the clauses' labels match the single given label of the\\
& & KN sentence.
\end{tabular}

\smallskip
\noindent However, under a {\em multi-label prediction} setting, when using clauses as prediction unit, ANY and ALL would be too extreme to give a fair estimate of the prediction quality. We therefore introduce a third variant, UNION, as follows.

\medskip\noindent
\begin{tabular}{lll}
\textbf{UNION} & ~TPs:~ & the {\em union} of the clauses' labels that match the given set of\\
& & labels of the KN sentence.\\
& ~FPs:~ & the {\em union} of the clauses' labels that do not match the given\\
& & set of labels of the KN sentence.
\end{tabular}

\smallskip
\noindent That is, under a multi-label prediction setting (both when using sentences and clauses as prediction units), multiple TPs, FPs and FNs may be produced per KN sentence. FewRel, on the other hand, is a single-labeled dataset and provides property annotations at the fact level. There, ALL and ANY collapse into the same (standard) case, while UNION is not present at all. Based on the afore-defined variants of TPs, FPs and FNs, we then compute {\em precision} (P), {\em recall} (R) and F1 in the standard way for both KN and FewRel.

\begin{table}
\caption{The performance of all our approaches using KnowledgeNet.}
\label{tab:ourKNresults}
\centering
\begin{tabular}{l||c}
\hline
\textbf{Method}  &P \enspace R \enspace F1\\ 

\hline
\hline

Human & 0.88 \enspace 0.88 \enspace 0.88  \\
Diffbot Joint Model & 0.81 \enspace 0.81 \enspace 0.81  \\
KnowledgeNet Baseline 5 (BERT) & 0.67 \enspace 0.69 \enspace 0.68  \\

\hline

Clauses + BERT (ALL) & 0.86 \enspace 0.86 \enspace 0.86  \\

Clauses + BERT (ANY) & 0.90  \enspace 0.92 \enspace 0.91 \\ 

Clauses + BERT (UNION) & 0.89 \enspace 0.89 \enspace 0.89 \\


Clauses + distillBERT (ALL) & 0.86 \enspace 0.86 \enspace 0.86 \\

Clauses + distillBERT (ANY) & \textbf{0.92} \enspace \textbf{0.92} \enspace \textbf{0.92} \\

Clauses + distillBERT (UNION) & 0.91 \enspace 0.91\enspace 0.91  \\


Clauses + feature-based-BERT (ALL) & 0.86 \enspace 0.74 \enspace 0.79 \\

Clauses + feature-based-BERT (ANY) & 0.91 \enspace 0.91 \enspace 0.91 \\

Clauses + feature-based-BERT (UNION)& 0.91 \enspace 0.87 \enspace 0.89  \\


Clauses + SETFIT(ANY) & 0.85 \enspace 0.83 \enspace 0.84  \\

\hline

Mixed + BERT (ALL) & 0.87 \enspace 0.75 \enspace 0.80  \\

Mixed + BERT (ANY) & 0.93  \enspace 0.84 \enspace 0.89 \\

Mixed + BERT (UNION) & \textbf{0.91} \enspace \textbf{0.93} \enspace \textbf{0.92} \\


Mixed + distillBERT (ALL) & 0.85 \enspace 0.70 \enspace 0.77 \\

Mixed + distillBERT (ANY) & 0.91 \enspace 0.80 \enspace 0.85 \\

Mixed + distillBERT (UNION) & 0.90 \enspace 0.92 \enspace 0.91  \\


Mixed + feature-based-BERT (ALL) & 0.85 \enspace 0.69 \enspace 0.76 \\

Mixed + feature-based-BERT (ANY) & 0.85 \enspace 0.83 \enspace 0.84 \\

Mixed + feature-based-BERT (UNION)& 0.88 \enspace 0.83 \enspace 0.85  \\


Mixed + SETFIT (ANY) & 0.82 \enspace 0.77 \enspace 0.79  \\

\hline


Sentences + BERT & 0.86 \enspace 0.78 \enspace 0.82  \\


Sentences +  distillBERT  &  \textbf{0.87} \enspace \textbf{0.79} \enspace \textbf{0.83} \\

\hline

Clauses + Word2Vec (ALL) & 0.71 \enspace 0.62 \enspace 0.66  \\

Clauses + Word2Vec (ANY) & 0.77 \enspace 0.58 \enspace 0.67 \\

Clauses + Word2Vec (UNION) & \textbf{0.83} \enspace \textbf{0.66} \enspace \textbf{0.66}
\\


\hline
\end{tabular}
\end{table}

\subsection{Results}
\label{sec:results}
We now present the results of the seven baseline approaches outlined in Section \ref{sec:baselines}. We performed detailed experiments to demonstrate the effectiveness of our method and show how OpenIE improves RE. We tested multiple pre-trained LMs and report the best results in Tables \ref{tab:ourKNresults} and \ref{tab:ourFRresults}. For all the baselines except Baseline 3 and 7, we consider candidate clauses when their subject and object text spans overlap with the text spans of the subject and object of the ground truth relations.   
For KN\footnote{\url{https://github.com/diffbot/knowledge-net}}, the results are averaged after performing a 4-fold cross-validation on the 4 folders into which it is divided by default. 
For FewRel\footnote{\url{http://zhuhao.me/fewrel}}, we averaged over 10 runs with random splits (each by dividing the dataset in 75\% for training and 25\% for testing purposes) to shuffle as much as possible the data and have significant changes in the distribution of the text during training and testing time. For further details on the pre-trained BERT models, we refer to \cite{Devlin:2018}.

\begin{table}
\begin{minipage}{\textwidth}
\caption{The performance of all our approaches using FewRel\protect\footnote{Here, for SETFIT, a subset of 15 relations is used similar to \cite{setfit}}.}
\label{tab:ourFRresults}
\centering
\begin{tabular}{l||c}
\hline
\textbf{Method}
&P \enspace R \enspace F1 \\
\hline
\hline
ERNIE & 0.88 \enspace 0.88 \enspace 0.88   \\
DeepEx & --- \enspace --- \enspace 0.48   \\
\hline

Clauses + BERT & \textbf{0.71} \enspace \textbf{0.71} \enspace \textbf{0.71}   \\
Clauses + distillBERT & 0.68 \enspace 0.68 \enspace 0.68 \\
Clauses + SETFIT & 0.68 \enspace 0.68 \enspace 0.68  \\
Clauses + roBERTa & 0.68 \enspace 0.68 \enspace 0.68 \\
Clauses + distillroBERTa & 0.66 \enspace 0.67 \enspace 0.66 \\
Clauses + feature-based-BERT & 0.75 \enspace 0.59 \enspace 0.66 \\
Clauses + alBERT & 0.65 \enspace 0.66 \enspace 0.65  \\

\hline

Mixed + BERT & \textbf{0.66} \enspace \textbf{0.67} \enspace \textbf{0.66}   \\
Mixed + distillBERT & 0.65 \enspace 0.66 \enspace 0.65 \\
Mixed + roBERTa & 0.65 \enspace 0.67 \enspace 0.65 \\
Mixed + distillroBERTa & 0.65 \enspace 0.67 \enspace 0.65 \\
Mixed + SETFIT & 0.66 \enspace 0.64 \enspace 0.65  \\
Mixed + alBERT & 0.62 \enspace 0.63 \enspace 0.62  \\
Mixed + feature-based-BERT & 0.64 \enspace 0.59 \enspace 0.61 \\

\hline 

Sentences + BERT & \textbf{0.65} \enspace \textbf{0.66} \enspace \textbf{0.65}   \\
Sentences +  roBERTa & 0.65 \enspace 0.66 \enspace 0.65 \\
Sentences +  distillroBERTa & 0.65 \enspace 0.66 \enspace 0.65 \\
Sentences +  alBERT & 0.63 \enspace 0.65 \enspace 0.64  \\
Sentences + distillBERT & 0.64 \enspace 0.65 \enspace 0.64 \\
Sentences +  SETFIT & 0.64 \enspace 0.63 \enspace 0.63  \\

\hline

Clauses + Word2Vec & 0.61 \enspace 0.52 \enspace 0.56  \\

\hline
\end{tabular}
\end{minipage}
\end{table}

\smallskip
\noindent\textbf{KN Results.~} We motivated our choice for the LMs in Section \ref{sub:bert}, however, the experimental results do not suggest a clear suitability of a specific model for all RE settings. We notice that BERT and distillBERT performed best on KN, while RoBERTa and SETFIT were also useful in some settings applied to FewRel. 
For KN, our best baseline (Baseline 1, Clauses + distillBERT) significantly outperforms the previous work (Diffbot Joint Model and KN Baseline 5
, reported on top of Table \ref{tab:ourKNresults}). The most important improvements (also in comparison to our other baselines) are due to (1) using clauses as a unit of prediction, (2) incorporating clauses during fine-tuning, and (3) allowing any of the OpenIE clauses to match the single KN label (as described in the ANY evaluation mode).

\smallskip
\noindent\textbf{FewRel Results.~} For FewRel, we compare our results against Matching the Blanks (MTB) \cite{SoaresFLK19}, ERNIE \cite{ERNIE} and DeepEx \cite{DeepEx}. Being the board leader on FewRel (with an accuracy of $93.86$), Matching the Blanks classifies the relations relying solely on the text input. It however employs additional entity markers, which we deliberately omit in favor of taking advantage of the OpenIE-based sentence decomposition and the LMs ability to interpret the arguments. While our strategy proves effective for KN, explicit entity markers may still be lacking for FewRel which represents a much more fine-grained set of 100 relations (compared to the 15 relations of KN).
ERNIE is different from MTB and our system in that it uses knowledge graphs to enrich pre-trained LM. It shows good performance on FewRel (with an F1-score of $88.32$), but the robustness of the system may be questioned due to inherent incompleteness of the knowledge graphs which typically limits the system's ability to generalize. We, on the other hand, want to demonstrate how a fast and simple approach can be successful even on such a competitive dataset while not suffering from unseen relational components.
DeepEx offers an interesting comparative scenario because it formulates the RE task as an extension to OpenIE. While DeepEx outscores many state of the art OpenIE systems, we outperform it on the task of RE by large margin, including the few-shot setting. We attribute this result to the way OpenIE clauses are translated into relations: DeepEx essentially maps relational phrases from clauses to a knowledge graph property label or its aliases but does not take the signal from the entire clause into account. We notice that this approach roughly corresponds to our context-free ``Clauses + Word2Vec'' baseline which generally achieves weaker results on both datasets (especially compromising the recall), as there seem to be a plethora of relation types, specially in FewRel, that cannot be captured well just by considering the clauses' verbal phrases.

\smallskip
\noindent\textbf{Few-Shot Results.~} Figure \ref{fig:fewshot} shows the best performing models in FewShot setting. We notice that for KN, 8 samples are sufficient for feature-based BERT to achieve about 85\% F1-score. The other two models require many more samples yet do not reach the same result. On the contrary, all the tree models demonstrate similar behaviour on FewRel data. BERT has a slight advantage, however, it needs at least 30 samples to achieve above 50\% F1-score. We hypothesize that the overall number and diversity of relations in FewRel contribute to this phenomenon.

\smallskip
\noindent\textbf{Summary.~}
To summarize, we clearly notice that the best performances are obtained on the baselines where we use OpenIE extractions for inference and fine-tuning the LMs, which motivates our choice and design for the experiments. Our experiments confirm that decomposing a sentence into clauses, which are designed to express a compact, more ``focused'' unit of information, allows us to distill various aspects of a sentence's meaning and to represent it as (a set of) labeled relation instances in a concise manner.

\begin{figure}
    \centering
    \includegraphics[width=1\textwidth]{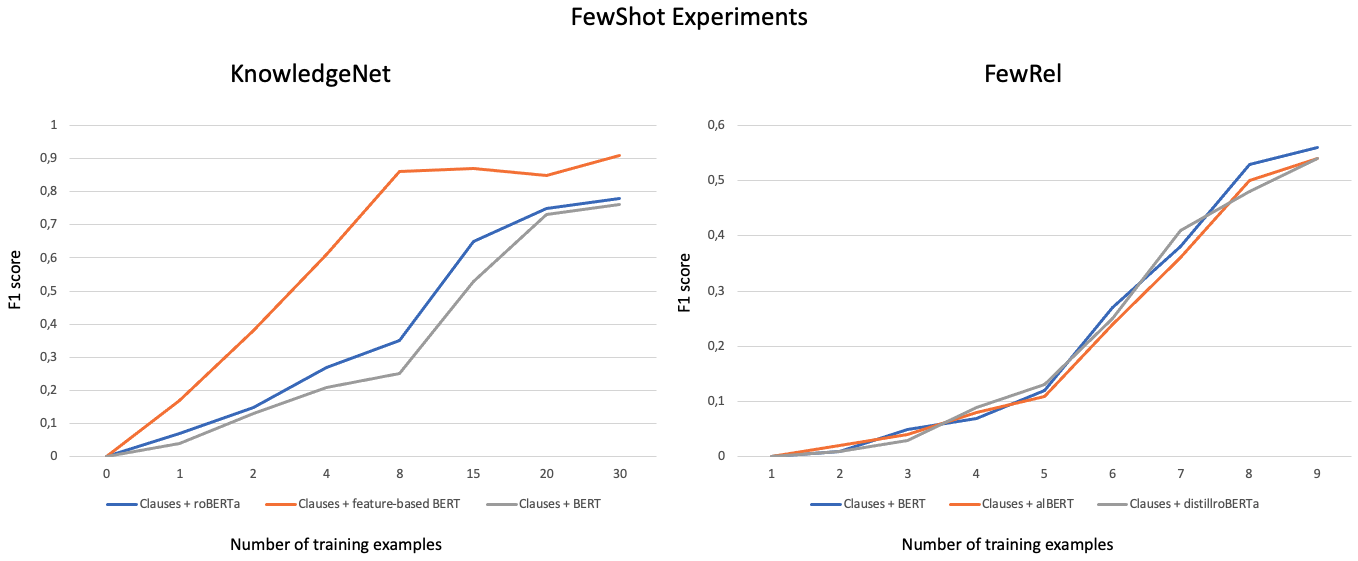}
    \caption{Change in the F1 score in a
 FewShot setting.}
    \label{fig:fewshot}
\end{figure}

\section{Conclusions}
\label{sec:conclusion}
We proposed and evaluated a variety of simple and direct strategies to combine OpenIE with Language Models for the task of Relation Extraction. We explored how OpenIE may serve as an intermediate way of extracting concise factual information from natural-language input sentences, and we combined the obtained clauses with both context-free and contextual LMs, as both are widely used in research and known for their strengths and weaknesses. For our experiments, we utilized the KnowledgeNet dataset with 15 properties as well as the well-known FewRel dataset containing 100 relations. Both datasets use Wikidata as underlying KB, which constitutes a valuable resource as demonstrated by the increased number of scientific publications and applications both in academia and industry in recent years. We presented detailed experiments on Word2Vec, BERT, RoBERTa, AlBERT, SETFIT and their further distilled versions with a range of baselines that achieve up to 92\% and 71\% of F1 score for KnowledgeNet and FewRel, respectively. In the future, we intend to apply our work also toward various downstreams tasks such as sentiment analysis, question answering, knowledge base population, and further knowledge graph aspects.

\subsubsection*{Acknowledgements.}
We thank Matteo Cannaviccio, co-author of KnowledgNet and engineer at Diffbot, for the fruitful discussion and the precious insights. 


%
%
%

\end{document}